\documentclass[runningheads]{llncs}
\usepackage[T1]{fontenc}
\usepackage{graphicx}
\usepackage[american]{babel}
\usepackage{tipa}
\usepackage{mathtools} 
\usepackage{booktabs} 
\usepackage{tikz} 
\usepackage{adjustbox}[2019/01/04]

\usepackage{amsmath,amsfonts,bm}

\def\eqref#1{equation~\ref{#1}}

\def\1{\bm{1}}

\DeclareMathAlphabet{\mathsfit}{\encodingdefault}{\sfdefault}{m}{sl}
\SetMathAlphabet{\mathsfit}{bold}{\encodingdefault}{\sfdefault}{bx}{n}

\renewcommand{\v}[1]{{\boldsymbol{\mathbf{#1}}}}
\newcommand\mask{m}
\newcommand{\pref}{\tilde{p}}

\begin{document}
%
\title{Object-based active inference}
%
%
\author{Ruben S. van Bergen\inst{1}\orcidID{0000-0001-9216-6531} \and
Pablo L. Lanillos\inst{1}\orcidID{0000-0001-9154-0798}}
\author{Ruben S. van Bergen \and Pablo L. Lanillos}

\authorrunning{van Bergen \& Lanillos}
%
\institute{Department of Artificial Intelligence \& \\ Donders Institute for Brain, Cognition \& Behavior \\ Radboud University, Nijmegen, the Netherlands \\ \email{\{ruben.vanbergen, pablo.lanillos\}@donders.ru.nl}}


\maketitle 
\begin{abstract}
The world consists of objects: distinct entities possessing independent properties and dynamics. For agents to interact with the world intelligently, they must translate sensory inputs into the bound-together features that describe each object. These object-based representations form a natural basis for planning behavior. Active inference (AIF) is an influential unifying account of perception and action, but existing AIF models have not leveraged this important inductive bias. To remedy this, we introduce ‘object-based active inference’ (OBAI), marrying AIF with recent deep object-based neural networks. OBAI represents distinct objects with separate variational beliefs, and uses selective attention to route inputs to their corresponding object slots. Object representations are endowed with independent action-based dynamics. The dynamics and generative model are learned from experience with a simple environment (active multi-dSprites). We show that OBAI learns to correctly segment the action-perturbed objects from video input, and to manipulate these objects towards arbitrary goals. 
\end{abstract}

\section{Introduction}
\label{sec:intro}

Intelligent agents are not passive entities that observe the world and learn its causality. They learn the relationship of action and effect by interacting with the world, in order to fulfil their goals~\cite{lanillos2016yielding}. In higher-order intelligence, such as exhibited by primates, these interactions very often take place at the level of objects~\cite{kourtzi2011neural,Peters2021}. Whether picking a ripe fruit from a tree branch, kicking a football, or taking a drink from a glass of water; all require reasoning and planning in terms of objects. Objects, thus, are natural building blocks for representing the world and planning interactions with it. 

While there have been recent advances in unsupervised multi-object representation learning and inference~\cite{greff2019multi,locatello2020object}, to the best of the authors knowledge, no existing work has addressed how to leverage the resulting representations for generating actions. In addition, object perception itself could benefit from being placed in an active loop, as carefully selected actions could resolve ambiguity about object properties (including their segmentations - i.e., which inputs belong to which objects). Meanwhile, state-of-the-art behavior-based learning (control), such as model-free reinforcement learning~\cite{mnih2015human} uses complex encoding of high-dimensional pixel inputs without taking advantage of objects as an inductive bias (though see \cite{Veerapaneni2019entityRL,Watters2019cobra}.

To bridge the gap between these different lines of work, we here introduce `object-based active inference' (OBAI, pronounced /\textipa{@'beI}/), a new framework that combines deep, object-based neural networks~\cite{greff2019multi} and active inference~\cite{parr2022active,lanillos2021active}. Our proposed neural architecture functions like a Bayesian filter that iteratively refines perceptual representations. Through selective attention, sensory inputs are routed to high-level object modules (or \textit{slots}~\cite{locatello2020object}) that encode each object as a separated probability distribution, whose evolution over time is constrained by an internal model of action-dependent object dynamics. These object representations are highly compact and abstract, thus enabling efficient unrolling of possible futures in order to select optimal actions in a tractable manner. Furthermore, we introduce a closed-form procedure to learn preferences or goals in the network's latent space.

As a proof-of-concept, we evaluate our proposed framework on an active version of the multi-dSprites dataset, developed for this work (See Fig.~\ref{fig:benchmark-and-architecture}a). Our preliminary results show that OBAI is able to: $i$) learn to segment and represent objects $ii$) learn the action-dependent, object-based dynamics of the environment; and $iii$) plan in the latent space -- obviating the need to imagine detailed pixel-level outcomes in order to generate behavior. This work is a first step towards building more complex object-based active inference systems that can perform more cognitively challenging tasks on naturalistic input.

\section{Methods}
\label{sec:methods}



\subsection{Object-structured generative model}
\label{sec:methods:iai}
We extend the IODINE architecture proposed in \cite{greff2019multi} for object representation learning, to incorporate dynamics. Zablotskaia et al.~\cite{zablotskaia2020unsupervised} previously developed a similar extension to IODINE, in which object dynamics were modeled implicitly, through LSTM units operating one level below the latent-space representation. Here, we instead implement the dynamics directly in the latent space, and allow these dynamics to be influenced by actions on the part of the agent. 

\begin{figure}[t]
	\centering
    \includegraphics[width=0.9\textwidth]{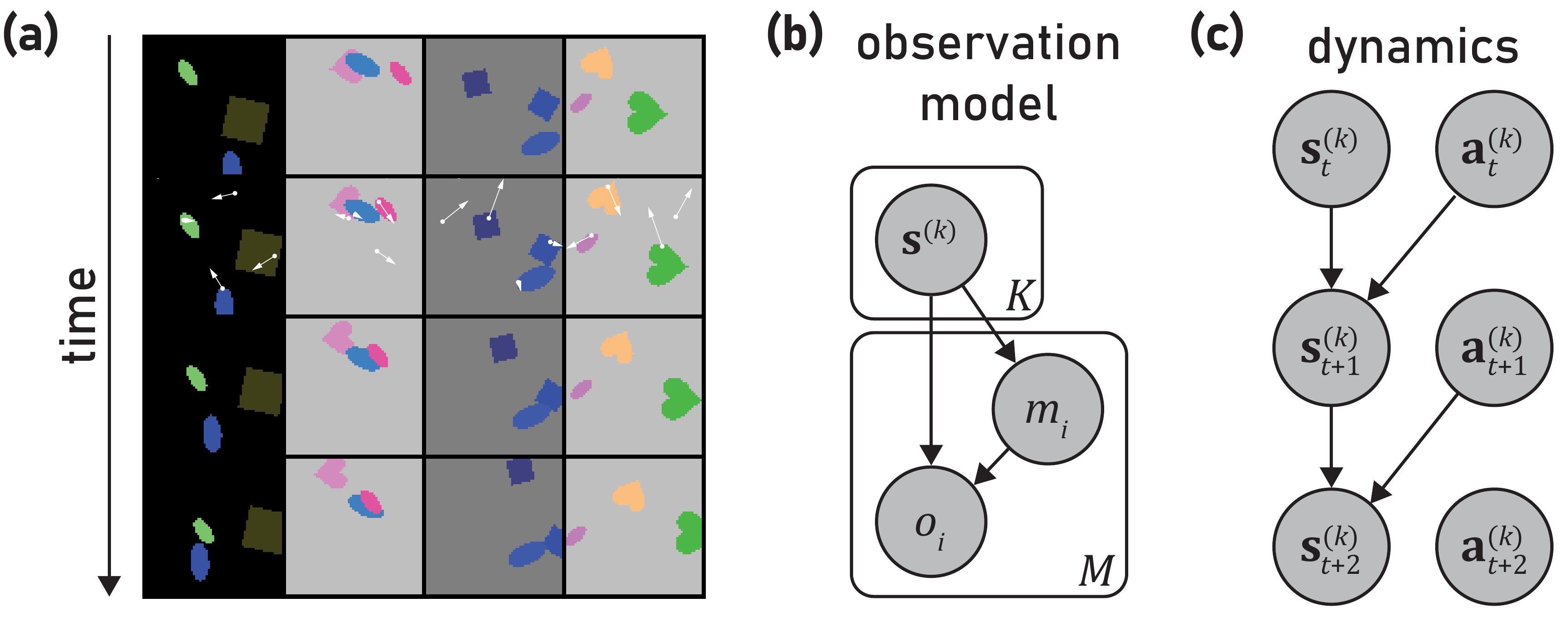}
	\caption{\textbf{Environment and generative model}. (a) Active multi-dSprites. Non-zero accelerations in action fields (in the 2nd frame) are indicated as white arrows originating at the accelerated grid location. (b) Object-structured generative model for a single image (time indices omitted here for clarity of exposition). (c) Dynamics model in state-space for a single object $k$, shown for three time points.}
	\label{fig:benchmark-and-architecture}
\end{figure}

Like IODINE, our framework relies on iterative amortized inference~\cite{Marino2018} (IAI) on an object-structured generative model. This model describes images of up to $K$ objects with a Normal mixture density (illustrated in Fig.~\ref{fig:benchmark-and-architecture}):
\small
\begin{equation}
    p(o_i|\{ \v{s}^{(k)}\}_{k\in1:K}, \mask_i) = \sum_k \left[ \mask_i=k  \right] \mathcal{N}\left(g_{i}(\v{s}^{(k)}), \sigma_o^2 \right)
\end{equation}
\normalsize
where $o_i$ is the value of the $i$-th image pixel, $\v{s}^{(k)}$ is the state of the $k$-th object, $g_i(\bullet)$ is a decoder function (implemented as a deep neural network (DNN)) that translates an object state to a predicted mean value at pixel $i$, $\sigma_o^2$ is the variability of pixels around their mean values and, crucially, $\mask_i$ is a categorical variable that indicates which object (out of a possible $K$ choices) pixel $i$ belongs to\footnote{Note the use of Iverson-bracket notation; the bracket term is binary and evaluates to 1 iff the expression inside the brackets is true.}. Note that the same decoder function is shared between objects. The pixel assignments themselves also depend on the object states:


\small
\begin{equation}
p(\mask_i| \{ \v{s}^{(k)} \}_{k\in1:K}) = \text{Cat}\left(\text{Softmax}\left(\{\pi_i( \v{s}^{(k)} )\}_{k\in1:K}\right)\right)
\end{equation}
\normalsize
where $\pi_i(\bullet)$ is another DNN that maps an object state to a log-probability at pixel $i$, which (up to a constant of addition) defines the probability that the pixel belongs to that object. Marginalized over the assignment probabilities, the pixel likelihoods are given by:
\small
\begin{gather}
    p(o_i|\{ \v{s}^{(k)}\}_{k\in1:K}) = \sum_k \hat{\mask}_{ik}  \mathcal{N}\left(g_{i}(\v{s}^{(k)}), \sigma_o^2 \right)\\
    \hat{\mask}_{ik} = p(\mask_i=k| \{ \v{s}^{(k)} \}_{k\in1:K}) 
\end{gather}
\normalsize
During inference, the soft pixel assignments $\{\hat{\mask}_{ik}\}$ introduce dynamics akin to selective attention, as each object slot is increasingly able to focus on those pixels that are relevant to that object.

\subsection{Incorporating action-dependent dynamics}
So far, this formulation is identical to the generative model in IODINE. We now extend this with an action-based dynamics model. We want to endow objects with (approximately) linear dynamics, and to allow actions that accelerate the objects. First, we redefine the state of an object at time point $t$ in generalized coordinates, i.e. $\v{s}^\dagger_t=\begin{bmatrix} \v{s}_t \\ \v{s}'_t \end{bmatrix}$, where $\v{s}'$ refers to the first-order derivative of the state. The action-dependent state dynamics are then given by:
\small
\begin{align}
    \v{s}_{t}^{\prime(k)} &= \v{s}_{t-1}^{\prime(k)} + \v{D}\v{a}_{t-1}^{(k)} + \sigma_s\v{\epsilon}_1  \label{eq:dynamics1} \\
    \v{s}_{t}^{(k)} &= \v{s}_{t-1}^{(k)} + \v{s}_{t}^{\prime(k)} + \sigma_s\v{\epsilon}_2 \label{eq:dynamics2}
\end{align}
\normalsize
where $\v{a}_t^{(k)}$ is the action on object $k$ at time $t$. This action is a 2-D vector that specifies the acceleration on the object in pixel coordinates. Multiplication by \v{D} (which is learned during training) transforms the pixel-space acceleration to its effect in the latent space\footnote{Since the network will be trained unsupervised, we do not know in advance the nature of the latent space representation that will emerge. In particular, we do not know in what format (or even if) the network will come to represent the positions of the objects.}. Equations \ref{eq:dynamics1}-\ref{eq:dynamics2} thus define the object dynamics model $p(\v{s}^{\dagger(k)}_{t}|\v{s}^{\dagger(k)}_{t-1}, \v{a}^{(k)}_{t-1})$. 


We established that $\v{a}_t^{(k)}$ is the action on object $k$ in the model at a given time. However, note that the correspondence between objects represented by the model, and the true objects in the (simulated) environment, is unknown.\footnote{In particular, since the representation across objects slots in the network is permutation-invariant, their order is arbitrary -- just as the order in the memory arrays that specify the environment is also arbitrary. Thus, the object-actions, as represented in the model, cannot be unambiguously mapped to objects in the environment that the agent interacts with. This problem is exacerbated if the network has not inferred the object properties and segmentations with perfect accuracy or certainty, and thus cannot accurately or unambiguously refer to a true object in the environment.} To solve this correspondence problem, we introduce the idea of \textit{action fields}. An action field $\v{\Psi}=[\v{\psi}_1, ..., \v{\psi}_M]^T$ is an $[M \times 2]$ matrix (with $M$ the number of pixels in an image or video frame), such that the $i$-th row in this matrix ($\v{\psi}_i$) specifies the (x,y)-acceleration applied at pixel $i$. In principle, a different acceleration can be applied at each pixel coordinate (in practice, we apply accelerations sparsely). These pixel-wise accelerations affect objects through the rule that each object receives the sum of all accelerations that occur at its visible pixels:
\small
\begin{equation} \label{eq:actionfield}
    \v{a}_t^{(k)} = \sum_i [\mask_i=k]\v{\psi}_i + \sigma_\psi \epsilon_3
\end{equation}
\normalsize
where we include a small amount of Normally distributed noise in order to make this relationship amenable to variational inference. This definition of actions in pixel-space is unambiguous and allows the model to interact with the environment. 

\subsection{Inference}
On the generative model laid out in the previous section, we perform \textit{iterative amortized inference} (IAI). IAI generalizes variational autoencoders (VAEs), which perform inference in a single feedforward pass, to architectures which use several iterations (implemented in a recurrent network) to minimize the Evidence Lower Bound (ELBO). As in VAEs, the final result is a set of variational beliefs in the latent space of the network. In our case, this amounts to inferring $q(\{\v{s}^{\dagger(k)}, \v{a}^{(k)}\}_{k\in1:K})$. We choose these beliefs to be independent Normal distributions.
Inference and learning both minimize the following ELBO loss:
\small
\begin{multline} \label{eq:elbo} 
    \mathcal{L} = -\sum_{t=0}^{T} \Bigg[ \mathcal{H}\left(q\left(\{\v{s}^{\dagger(k)}_t, \v{a}^{(k)}_t\}\right) \right) +  E_{q(\{\v{s}^{(k)}_t\})}[\log p(\v{o}_t|\{ \v{s}^{(k)}_t\})]  \\
    + \sum_k E_{q(\v{a}^{(k)}_t)}[\log p(\v{a}^{(k)}_t|\v{\Psi}_t)]  + \sum_k E_{q\left(\v{s}^{\dagger(k)}_{t}, \v{s}^{\dagger(k)}_{t-1}, \v{a}^{(k)}_{t-1}\right)}[\log p(\v{s}^{\dagger(k)}_{t}|\v{s}^{\dagger(k)}_{t-1}, \v{a}^{(k)}_{t-1})]  \Bigg] 
\end{multline} 
\normalsize
for some time horizon $T$. Note that for $t=0$, we define $p(\v{s}^{\dagger(k)}_{t}|\v{s}^{\dagger(k)}_{t-1}, \v{a}^{(k)}_{t-1})=p(\v{s}^{\dagger(k)})=\prod_{jk}\mathcal{N}(s^{\dagger(k)}_j; 0,1)$, i.e. a fixed standard-Normal prior. To compute $E_{q(\v{a}^{(k)}_t)}[\log p(\v{a}^{(k)}_t|\v{\Psi}_t)]$, we employ a sampling procedure, described in Appendix~\ref{sec:samplingtrick}.


The IAI architecture consists of a decoder module that implements the generative model, and a refinement module which outputs updates to the parameters $\v{\lambda}$ of the variational beliefs. Mirroring the decoder, the refinement module consists of $K$ copies of the same network (sharing the same parameters), such that refinement network $k$ outputs updates to $\v{\lambda}^{(k)}$. Network architectures for the decoder and refinement modules are detailed in Appendix~\ref{sec:network-architectures}. To perform inference across multiple video frames, we simply copy the refinement and decoder networks across frames as well as object slots. Importantly, as in \cite{greff2019multi}, each refinement network instance also receives as input a stochastic estimate of the current gradient $\nabla_{\v{\lambda}_t^{(k)}}  \mathcal{L}$. Since the ELBO loss includes a temporal dependence term between time points, the inference dynamics in the network are automatically coupled between video frames, constraining the variational beliefs to be consistent with the dynamics model. To infer $q(\{\v{a}^{(k)}\}_{k\in1:K})$, we employ a separate (small) refinement network (again copied across objects slots; details in Appendix~\ref{sec:action-inference-module}).

\subsection{Task \& training}
We apply OBAI to a simple synthetic environment, developed for this work, which we term \textit{active-dSprites}. This environment was created by re-engineering the original dSprites shapes~\cite{dsprites17}, to allow these objects to be translated at will and by non-integer pixel offsets. The environment simulates these shapes moving along linear trajectories that can be perturbed through the action fields we introduced above. In the current work, OBAI was trained on pre-generated experience with this environment, with action fields sampled arbitrarily (i.e. not based on any intent on the part of the agent). 

Specifically, we generated video sequences (4 frames, 64$\times$64 pixels) from the active-dSprites environment, with 3 objects per video, in which we applied an action field only at a single time point (in the 2nd frame). Action fields were sparsely sampled such that (whenever possible) every object received exactly one non-zero acceleration at one of its pixels. Exactly one background pixel also received a non-zero acceleration, to encourage the model to learn not to assign background pixels to the segmentation masks of foreground objects. In practice, the appearance of the background was unaffected by these actions (conceptually, the background can be thought of as an infinitely large plane extending outside the image frame, and thus shifting it by any amount will not change its visual appearance in the image). 


OBAI was trained on 50,000 pre-generated video sequences, to minimize the ELBO loss from equation \ref{eq:elbo}. This loss was augmented to include the losses at intermediate iterations of the inference procedure, and this composite loss was backpropagated through time to compute parameter gradients. More details about the environment and training procedure can be found in Appendices~\ref{sec:active-dsprites}~\&~\ref{sec:training-procedure}.

\subsubsection{Learning goals in the latent space}\label{sec:methods:pref}
The active-dSprites environment was conceived to support cognitive tasks that require object-based reasoning. A natural task objective in active-dSprites is to move a certain object to a designated location. This type of objective is simple in and of itself, but the rules that determine which object must be moved where can be arbitrarily complex. For now, we restrict ourselves to the simple objective of moving all objects in a scene to a fixed location, and focus on how to encode this objective. We follow previous Active Inference work (e.g. \cite{Friston2015,Sajid2021demystified}) in conceptualizing goals as a preference distribution $\pref$. However, rather than defining this preference to be over observations, as is common (though see \cite{Friston2019particular,da2020active,Millidge2020whence}), we instead opt to define it over latent states, i.e. $\pref(\{\v{s}^{\dagger(k)}\})$, which simplifies action selection (a full discussion of the merits of this choice is outside the scope of this paper). 

Assuming that we can define a preference over the true state of the environment, $\v{s}_\text{true}$ (e.g. the ground-truth object positions), the preference distribution in latent space can be obtained through the following importance-sampling procedure:
\small
\begin{gather}
    \pref(\v{s}) \propto \sum_j p(\v{s}|\v{o}^*_j)u_j \approx \sum_j q(\v{s}|\v{o}^*_j)u_j \\
    \v{o}^*_j \sim p(\v{o} | \v{s}_{\text{true}_j}^*), \quad
    u_j = \pref( \v{s}_{\text{true}_j}^*), \quad
    \v{s}_{\text{true}_j}^* \sim p( \v{s}_\text{true}) \propto \text{Constant}
\end{gather}\normalsize
This allows the latent-space preference to be estimated in closed form from a set of training examples, constructed by sampling true states uniformly from the environment and rendering videos from these states. Inference is performed on the resulting videos, and the latent-state preference is computed as the importance-weighted average of the inferred state-beliefs (alternatively, we can sample states directly from $\pref( \v{s}_{\text{true}}^*)$, and let the importance weights drop out of the equation). In particular, if $\pref( \v{s}_{\text{true}}^*)$ is Normal, then the preference in the latent space is also Normal:
\small
\begin{gather}
    q(\v{s}|\v{o}^*_j) = \mathcal{N}\left(\v{\mu}(\v{o}^*_j), \v{\sigma}(\v{o}^{*2}_j)\right), \quad \pref(\v{s}) = \mathcal{N}(\tilde{\v{\mu}}, \tilde{\v{\sigma}}^2) \\
    \tilde{\v{\mu}} = \frac{1}{\sum_j u_j} \sum_j u_j \v{\mu}(\v{o}_j^*), \quad
    \tilde{\v{\sigma}} = \sqrt{\frac{1}{\sum_j u_j}  \sum_j u_j \left((\tilde{\v{\mu}} - \v{\mu}(\v{o}^*_j))^2    +  \v{\sigma}(\v{o}^{*}_j)^2 \right) }
\end{gather} \normalsize

\subsubsection{Planning actions}
OBAI can plan actions aimed at bringing the environment more closely in line with its learned preferences. Specifically, we choose actions that minimize the Free Energy of the Expected Future (FEEF) \cite{Millidge2020whence}. When preferences are defined with respect to latent states, the FEEF of a policy (action sequence) $\pi = \left\{[\v{a}^{(k)}_1, \v{a}^{(k)}_2, \hdots, \v{a}^{(k)}_T ] \right\}_{k\in 1:K}$ is given by:
\begin{gather}
    \mathcal{G}(\pi) = \sum_{\tau=1}^{T} \sum_k D_{KL}\left(q(\v{s}_\tau^{(k)}|\pi) || \tilde{p}(\v{s}) \right)     
\end{gather}
where $q(\v{s}_\tau^{(k)}|\pi)$ is the policy-conditioned variational prior, obtained by propagating the most recent state beliefs through the dynamics model by the requisite number of time steps.

Given this objective, the optimal policy can be calculated in closed form for arbitrary planning horizons. In this work, as a first proof-of-principle, we only consider greedy, one-step-ahead planning. In this case, the optimal "policy" (single action per object) is given by:
\small
\begin{equation}
    \hat{\v{a}}^{(k)} = (\v{D}^T\v{L}\v{D})^{-1}\v{D}^T\v{L}(\tilde{\v{\mu}}-\v{\mu}_{s}^{(k)})
\end{equation}
\normalsize
where $\v{L}=\text{diag}(\tilde{\v{\sigma}}^{-2})$, and $\v{\mu}_{s}^{(k)}$ is the mean of the current state belief for object $k$.


\section{Results}
\label{sec:results}

\subsection{Object segmentation and reconstruction in dynamic scenes}
\begin{figure}[t!]
	\centering
    \includegraphics[width=0.9\textwidth]{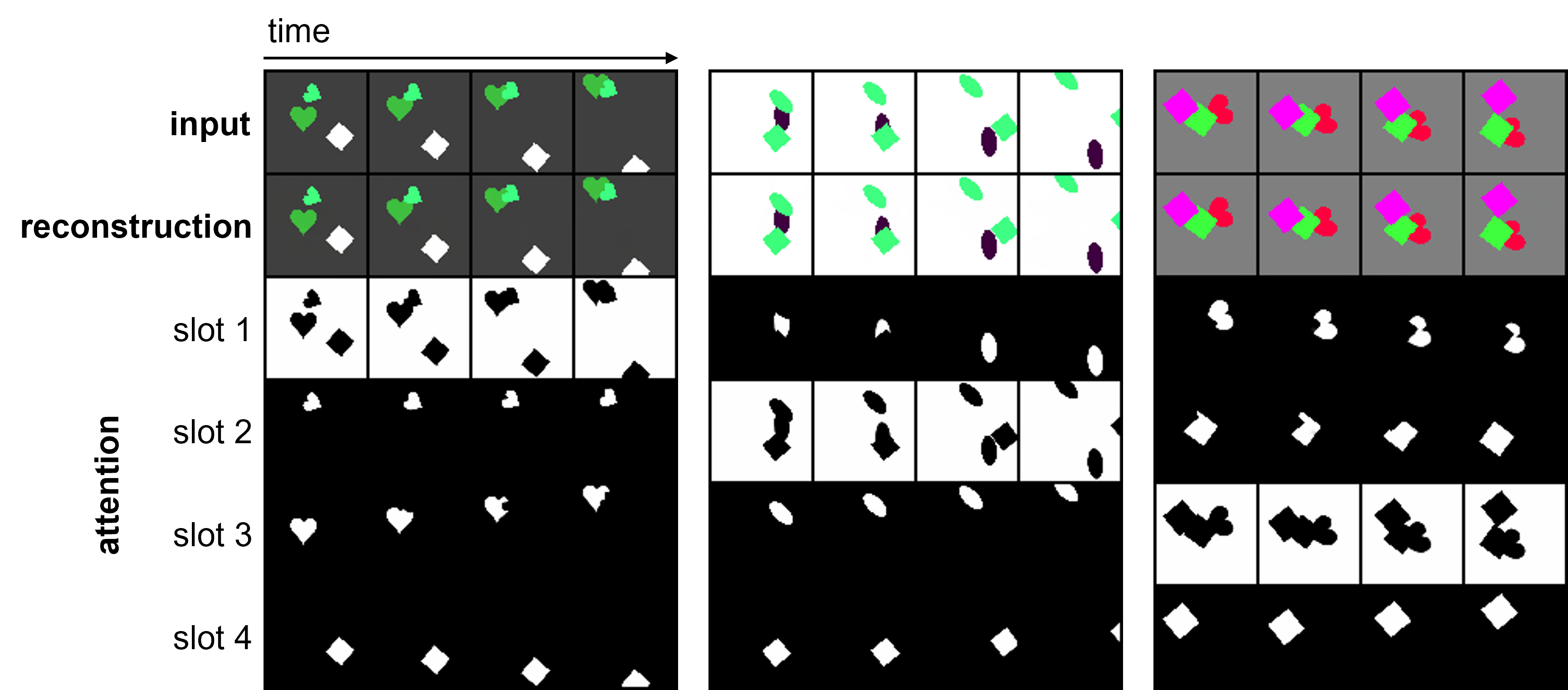}
	\caption{\textbf{Reconstruction and segmentation of videos of moving objects with actions}. Three instances of segmentation from the test set of videos. The masks shows how each slot attends to the three objects and the background. The action field in this experiment is randomly generated for each instance at the beginning of the simulation.}
	\label{fig:segmentation}
\end{figure}

We first evaluated OBAI on its inference and reconstruction capabilties, when presented with novel videos of moving objects (not seen during training). To evaluate this, we examined the quality of its object segmentations, and of the video frame reconstructions (Fig. \ref{fig:segmentation}). Segmentation quality was computed using the Adjusted Rand Index (ARI), as well as a modified version of this index that only considers (ground-truth) foreground pixels (FARI). Across a test set of 10,000 4-frame video sequences of 3 randomly sampled moving objects each, OBAI achieved an average ARI and FARI of 0.948 and 0.939, respectively (where 1 means perfect accuracy and 0 equals chance-level performance), and a MSE of $9.51\times10^{-4}$ (note that pixel values were in the range of $[0,1]$). For comparison, a re-implementation of IODINE, trained on 50,000 static images of 3 dSprite objects, achieved an ARI of 0.081, FARI of 0.856 and MSE of $1.63\times10^{-3}$ (on a test set of identically sampled static images). The very low ARI score reflects the fact that IODINE has no built-in incentive to assign background pixels to their own object slot. OBAI, on the other hand, has to account for the effects of actions being applied to the background, which must not affect the dynamics of the foreground objects. Thus, for OBAI to accurately model the dynamics in the training data, it must learn not to assign background pixels to the segmentation masks of foreground objects, lest and action might be placed on one of these spurious pixels. 

\subsection{Predicting the future state of objects}
\begin{figure}[t!]
	\centering
    \includegraphics[width=0.7\textwidth]{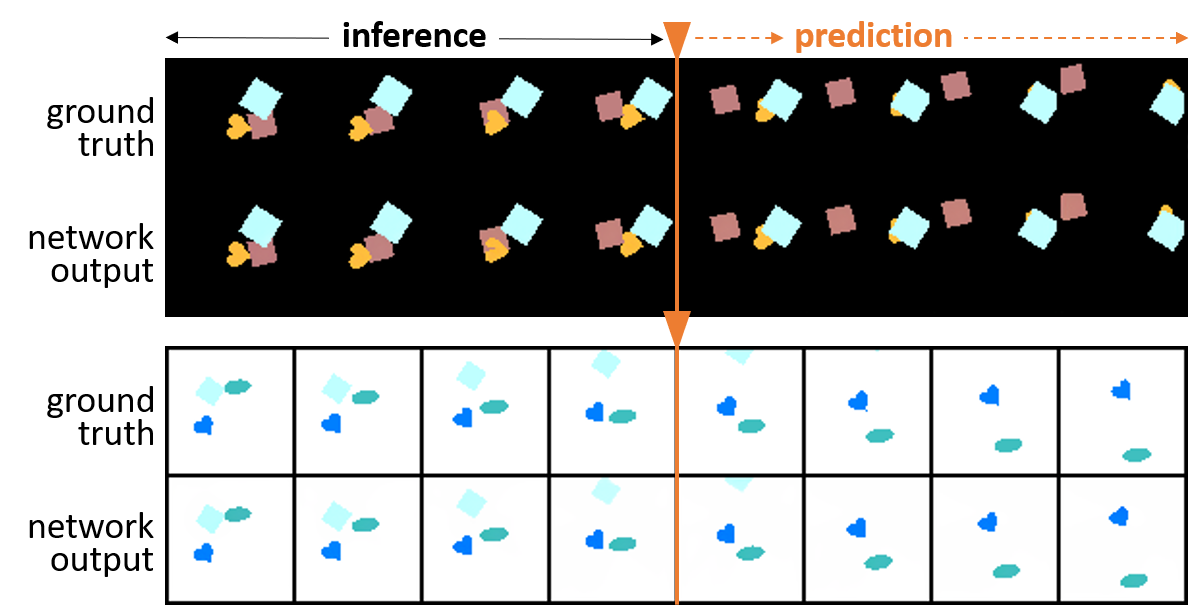}
	\caption{\textbf{Prediction of future video frames}. Two instances of prediction from the test set of videos. We let the network perform inference on 4 consecutive frames, and then predict the future. }
	\label{fig:prediction}
\end{figure}

An advantage of our approach is that the network can predict future states of the world at the level of objects using the learned state dynamics. Figure \ref{fig:prediction} shows three examples of the network predicting future video frames. The first 4 video frames are used by the network to infer the state. Afterwards, we extrapolate the inferred state and dynamics of the last observed video frame into the future, and decode the thus-predicted latent states into predicted video frames. These predictions are highly accurate, with a MSE of $4\times10^{-3}$.

\begin{table}[t!]
    \centering
    \begin{tabular}{lccl}
      \toprule 
      \bfseries  & \bfseries ARI ($\uparrow$)~~ & \bfseries  F-ARI ($\uparrow$)~~ & \bfseries MSE  ($\downarrow$) ~~ \\
      \midrule 
      IODINE (static) & $0.081$ & $0.856$ & $1.63 \times 10^{-3}$ \\
      OBAI (ours; 4 frames) & $0.948$ & $0.939$ & $9.51 \times 10^{-4}$ \\  
      \bottomrule 
      \addlinespace[6pt]
    \end{tabular}
    \caption{\textbf{Quantitative segmentation and reconstruction results.}}
    \label{tab:segment_recon_results}
\end{table}

\subsection{Goal-directed action planning}
\begin{figure}[b!]
	\centering
    \includegraphics[width=0.9\textwidth]{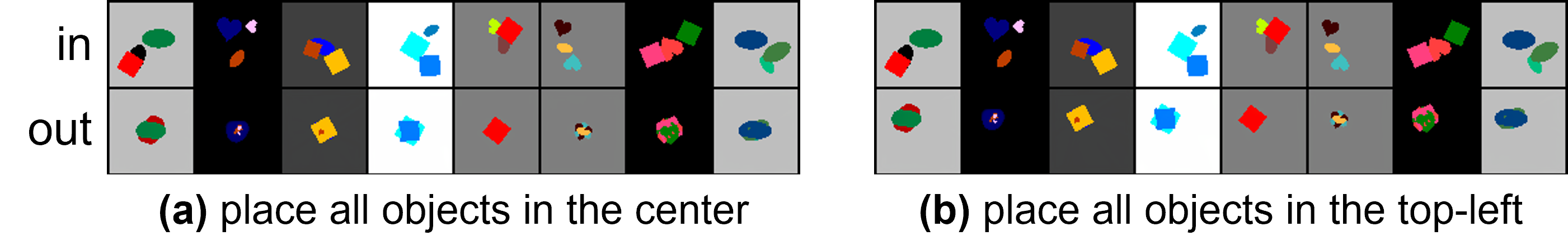}
	\caption{\textbf{Goal-directed action planning}. We give the network an arbitrary input image with three objects (in) and it infers the action that will move the state towards the learned preference, and imagines the resulting image (out). In (a), $\pref(\v{s}_\text{true})$ was biased towards the center of the image; in (b), it was biased more towards the top-left.}
	\label{fig:goal}
\end{figure}

Can OBAI learn and accomplish behavioral objectives? As a first foray into this question, we asked OBAI to learn fixed preference distributions defined in the true state-space of the environment, using the method described in section~\ref{sec:methods:pref}. Specifically, we placed a Gaussian preference distribution on the location of the object and had the network learn the corresponding preference in its latent space from a set of 10,000 videos (annotated with the requisite importance weights). We then presented the network with static images of dSprite objects in random locations, and asked it to "imagine" the action that would bring the state of the environment into alignment with the learned preference. Finally, we applied this imagined action to the latent state, and decoded the image that would result from this. As illustrated in Figure~\ref{fig:goal}, the network is reliably able to imagine actions that would accomplish the goals we wanted it to learn. 

\section{Conclusion}
\label{sec:discussion}
This work seeks to bridge an important gap in the field. On the one hand, computer vision research has developed object-based models, but these only perform passive inference. On the other hand, there is a wealth of research on behavioral learning (e.g. reinforcement learning and active inference), which has not generally leveraged objects as an inductive bias, built into the network architecture (cf. \cite{Veerapaneni2019entityRL,Watters2019cobra}). OBAI reconciles these two lines of research, by extending object-based visual inference models with action-based dynamics. We showed that OBAI can accurately track and predict the dynamics (and other properties) of simple objects whose movements are perturbed by actions -- an important prerequisite for an agent to plan its own actions. In addition, we presented an efficient method for internalizing goals as a preference distribution over latent states, and showed that the agent can infer the actions necessary to accomplish these goals, at the same abstract level of reasoning. While our results are preliminary, they are an important proof-of-concept, establishing the potential of our approach. In future work, we aim to scale OBAI to more naturalistic environments, and more cognitively demanding tasks. 

\section{Acknowledgements}
RSvB is supported by Human Brain Project Specific Grant Agreement 3 grant ID 643945539: "SPIKEFERENCE".

\bibliography{references.bib}






\newpage
\section*{\LARGE{Appendix}}
\renewcommand\thesection{\Alph{section}}
\setcounter{section}{0}
\section{Active-dSprites}\label{sec:active-dsprites}
Active-dSprites can be thought of as an "activated" version of the various multi-dSprites datasets that have been used in previous work on object-based visual inference (e.g.~\cite{greff2019multi,locatello2020object}). Not only does it include dynamics, but these dynamics can can be acted on by an agent. Thus, active-dSprites is an interactive environment, rather than a dataset. 

Objects in the active-dSprites environment are 2.5-D shapes (squares, ellipses and hearts): they have no depth dimension of their own, but can occlude each other within the depth dimension of the image. When an active-dSprites instance is intialized, object shapes, positions, sizes and colors are all sampled Uniformly at random. Initial velocities are drawn from a Normal distribution with mean 0 and standard deviation 4 (in units of pixels). Shape colors are sampled at discrete intervals spanning the full range of RGB-colors. Shapes are presented in random depth order against a solid background with a random grayscale color. Accelerations in action fields (at those locations that have been selected to incur a non-zero acceleration) are drawn from a Normal distribution with mean 0 and s.d. of 4. 

\section{Network architectures}\label{sec:network-architectures}
The OBAI architecture discussed in this paper consists of two separate IAI modules, each of which in turn contains a refinement and a decoder module. The first IAI module concerns the inference of the state beliefs $q(\{\v{s}^{\dagger(k)}\})$ -- we term this the \textit{state inference module}. The second IAI module infers the object action beliefs $q(\{\v{a}^{(k)}\})$, and we refer to this as the \textit{action inference module}.

We ran inference for a total of $F\times4$ iterations, where $F$ is the number of frames in the input. Inference initially concerns just the first video frame, and beliefs for this frame are initialized to $\v{\lambda}_0$, which is learned during training. After every 4 iterations, an additional frame is added to the inference window, and the beliefs for this new frame are initialized predictively, by extrapolating the object dynamics inferred up to that point. This procedure minimizes the risk that object representations are swapped between slots across frames, which can constitute a local minimum for the ELBO loss and leads to poor inference. We trained all IAI modules with $K$=4 object slots.

\subsection{State inference module}
This module used a latent dimension of 16. Note that, in the output of the refinement network, this number is doubled once as each latent belief is encoded by a mean and variance, and then doubled again as we represent (and infer) both the states and their first-order derivatives. In the decoder, the latent dimension is doubled only once, as the state derivatives do not enter into the reconstruction of a video frame. As in IODINE~\cite{greff2019multi}, we use a spatial broadcast decoder, meaning that the latent beliefs are copied along a spatial grid with the same dimensions as a video frame, and each latent vector is concatenated with the $(x,y)$ coordinate of its grid location, before passing through a stack of transposed convolution layers. Decoder and refinement network architectures are summarized in the tables below. The refinement network takes in 16 image-sized inputs, which are identical to those used in IODONE~\cite{greff2019multi}, except that we omit the leave-one-out likelihoods. Vector-sized inputs join the network after the convolutional stage (which processes only the image-sized inputs), and consist of the variational parameters and (stochastic estimates of) their gradients. 

\vspace{12pt}
\noindent\normalsize{\textbf{Decoder}} 
\\*\\*
\noindent\begin{tabular}{lccl}
  \toprule 
  \bfseries Type & \bfseries Size/\#Chan.~~ & \bfseries Act. func.~~ & \bfseries Comment \\
  \midrule 
  Input ($\v{\lambda}$) & 32 &  & \\ 
  Broadcast & 34 & & Appends coordinate channels \\
  Conv$^T$ $5 \times 5$ & 32 & ELU & \\
  Conv$^T$ $5 \times 5$ & 32 & ELU & \\
  Conv$^T$ $5 \times 5$ & 32 & ELU & \\
  Conv$^T$ $5 \times 5$ & 32 & ELU & \\
  Conv$^T$ $5 \times 5$ & 4 &  & Outputs RGB + mask\\
  \bottomrule 
  \addlinespace[6pt]
\end{tabular}

\vspace{12pt}
\noindent\normalsize{\textbf{Refinement network}} 
\\*\\*
\noindent\begin{tabular}{lccl}
  \toprule 
  \bfseries Type & \bfseries Size/\#Chan.~~ & \bfseries Act. func.~~ & \bfseries Comment \\
  \midrule 
  Linear & 64 & & \\ 
  LSTM & 128 & tanh & \\
  Concat $[..., \v{\lambda}, \nabla_\v{\lambda}\mathcal{L}]$ & 256 & & Appends vector-sized inputs \\
  Linear & 128 & ELU & \\
  Flatten & 800 & & \\
  Conv $5 \times 5$ & 32 & ELU & \\
  Conv $5 \times 5$ & 32 & ELU & \\
  Conv $5 \times 5$ & 32 & ELU & \\
  Inputs & 16 & & \\
  \bottomrule 
  \addlinespace[6pt]
\end{tabular}

\subsection{Action inference module}\label{sec:action-inference-module}
The action inference module does not incorporate a decoder network, as the quality of the action beliefs is computed by evaluating equation \ref{eq:actionfield} and plugging this into the ELBO loss from equation \ref{eq:elbo}. While this requires some additional tricks (see Appendix~\ref{sec:samplingtrick}), no neural network is required for this. This module does include a (shallow) refinement network, which is summarized in the table below. This network takes as input the current variational parameters $\v{\lambda_a}^{(k)}$ (2 means and 2 variances), their gradients, and the `expected object action', $\sum_i \hat{\mask}_{ik} \v{\psi}_i$. 

\vspace{12pt}
\noindent\normalsize{\textbf{Refinement network}} 
\\*\\*
\noindent\begin{tabular}{lccl}
  \toprule 
  \bfseries Type~~ & \bfseries Size/\#Chan.~~ & \bfseries Act. func.~~ & \bfseries Comment \\
  \midrule 
  Linear & 4 & & \\ 
  LSTM & 32 & tanh & \\
  Inputs & 10 & & \\
  \bottomrule 
  \addlinespace[6pt]
\end{tabular}

\section{Training procedure}\label{sec:training-procedure}
The above network architecture was trained on pre-generated experience with the active-dSprites environment, as described in the main text. The training set comprised 50,000 videos of 4 frames each. An additional validation set of 10,000 videos was constructed using the same environment parameters as the training set, but using a different random seed. Training was performed using the ADAM optimizer [REF] with default parameters and an initial learning rate of $3\times10^{-4}$. This learning rate was reduced automatically by a factor 3 whenever the validation loss had not decreased in the last 10 training epochs, down to a minimum learning rate of $3\times10^{-5}$. Training was performed with a batch size of 64 (16 $\times$ 4 GPUs), and was deemed to have converged after 245 epochs. 

\subsection{Modified ELBO loss}
OBAI optimizes an ELBO loss for both learning and inference. The basic form of this loss is given by equation~\ref{eq:elbo}. In practice, we modify this loss in two ways (similar to previous work, e.g.~\cite{greff2019multi}). First, we re-weight the reconstruction term in the ELBO loss as follows:
\begin{multline}
    \mathcal{L}_\beta = -\sum_{t=0}^{T} \Bigg[ \mathcal{H}\left(q\left(\{\v{s}^{\dagger(k)}_t, \v{a}^{(k)}_t\}\right) \right) +  \beta E_{q(\{\v{s}^{(k)}_t\})}[\log p(\v{o}_t|\{ \v{s}^{(k)}_t\})]  \\
    + \sum_k E_{q(\v{a}^{(k)}_t)}[\log p(\v{a}^{(k)}_t|\v{\Psi}_t)]  + \sum_k E_{q\left(\v{s}^{\dagger(k)}_{t}, \v{s}^{\dagger(k)}_{t-1}, \v{a}^{(k)}_{t-1}\right)}[\log p(\v{s}^{\dagger(k)}_{t}|\v{s}^{\dagger(k)}_{t-1}, \v{a}^{(k)}_{t-1})]  \Bigg]
\end{multline}
Second, we train the network to minimize not just the loss at the end of the inference iterations through the network, but a composite loss that also includes the loss after earlier iterations. Let $\mathcal{L}_\beta^{(n)}$ be the loss after $n$ inference iterations, then the composite loss is given by:
\begin{equation}
    \mathcal{L}_\text{comp}= \sum_{n=1}^{N_\text{iter}} \frac{n}{N_\text{iter}}\mathcal{L}_\beta^{(n)}
\end{equation}

\subsection{Hyperparameters}
OBAI includes a total of 4 hyperparameters: (1) the loss-reweighting coefficient $\beta$ (see above); (2) the variance of the pixels around their predicted values, $\sigma_o^2$; (3) the variance of the noise in the latent space dynamics, $\sigma_s^2$; and (4) the variance of the noise in the object actions, $\sigma_\psi^2$. The results described in the current work were achieved with the following settings:
\\ \\
\noindent\begin{tabular}{cc}
\toprule
\bfseries Param.~~ & \bfseries Value~~ \\
\midrule
$\beta$       & 5.0 \\
$\sigma_o$    & 0.3 \\
$\sigma_s$    & 0.1 \\
$\sigma_\psi$ & 0.3 \\
\bottomrule
\end{tabular}

\section{Computing $E_{q(\v{a}^{(k)})}[\log p(\v{a}^{(k)}|\v{\Psi})]$} \label{sec:samplingtrick}
The expectation under $q(\v{a}^{(k)})$ of $\log p(\v{a}^{(k)}|\v{\Psi})$, which appears in the ELBO loss (eq. \ref{eq:elbo}), cannot be computed in closed form, because the latter log probability requires us to marginalize over all possible configurations of the pixel-to-object assignments, and to do so inside of the logarithm. That is:
\begin{align}
    \log p(\v{a}^{(k)}|\v{\Psi}) &= \sum_\v{\mask} \log \left( p(\v{a}^{(k)}|\v{\Psi}, \v{\mask}) p(\v{\mask}|\{ \v{s}^{(k)} \}) \right) \\
    &= \log \left( E_{p(\v{\mask}|\{ \v{s}^{(k)} \})} [ p(\v{a}^{(k)}|\v{\Psi}, \v{\mask}) ] \right)
\end{align}
However, note that within the ELBO loss, we want to maximize the expected value of this quantity (as its negative appears in the ELBO, which we want to minimize). From Jensen's inequality, we have:
\begin{equation} \label{eq:jensen}
      E_{p(\v{\mask}|\{ \v{s}^{(k)} \})} [ \log p(\v{a}^{(k)}|\v{\Psi}, \v{\mask}) ] 
    \leq \log \left( E_{p(\v{\mask}|\{ \v{s}^{(k)} \})} [ p(\v{a}^{(k)}|\v{\Psi}, \v{\mask}) ] \right)
\end{equation}
Therefore, the l.h.s. of this equation provides a lower bound on the quantity we want to maximize. Thus, we can approximate our goal by maximizing this lower bound instead. This is convenient, because this lower bound, and its expectation under $q(\v{a}^{(k)})$ can be approximated through sampling:
\begin{gather}
     E_{q(\v{a}^{(k)})} \left[ E_{p(\v{\mask}|\{ \v{s}^{(k)} \})} [ \log p(\v{a}^{(k)}|\v{\Psi}, \v{\mask}) ] \right] \approx \frac{1}{N_\text{samples}} \sum_j  \log p(\v{a}^{(k)*}_j|\v{\Psi}, \v{\mask}^*_j) \\
     = \frac{1}{N_\text{samples}} \sum_j \log \mathcal{N}\left(\v{a}^{(k)*}_j; \sum_i \hat{\mask}^{*(i)}_{jk}\v{\psi}_i, \sigma_\psi^2\v{I}\right) \\
     \v{\hat{\mask}}^{*(i)}_j \sim  p(\mask_i|\{ \v{s}^{(k)} \}), \quad
     \v{a}^{(k)*}_j \sim q(\v{a}^{(k)}), \quad 
     \v{s}^{(k)*}_j \sim q(\v{s}^{(k)})
\end{gather}
where we slightly abuse notation in the sampling of the pixel assignments, as a vector is sampled from a distribution over a categorical variable. The reason this results in a vector is because this sampling step uses the Gumbel-Softmax trick~\cite{Jang2016gumbel}, which is a differentiable method for sampling categorical variables as "approximately one-hot" vectors. Thus, for every pixel $i$, we sample a vector $\v{\hat{\mask}}^{*(i)}_j$, such that the $k$-th entry of this vector, $\hat{\mask}^{*(i)}_{jk}$, denotes the "soft-binary" condition of whether pixel $i$ belongs to object $k$. In practice, we use $N_\text{samples}=1$, based on the intuition that this will still yield a good approximation over many training instances, and that we rely on the refinement network to learn to infer good beliefs. The Gumbel-Softmax sampling method depends on a temperature $\tau$, which we gradually reduce across training epochs, so that the samples gradually better approximate the ideal one-hot vectors. 

It is worth noting that, as the entropy of $p(\v{\mask}|\{ \v{s}^{(k)} \})$ decreases (i.e. as object slots "become more certain" about which pixels are theirs), the bound in equation \ref{eq:jensen} becomes tighter. In the limit as the entropy becomes 0, the network is perfectly certain about the pixel assignments, and so the distribution collapses to a point mass. The expectation then becomes trivial, and so the two sides of eq. \ref{eq:jensen} become equal. Sampling the pixel assignments is equally trivial in this case, as the distribution has collapsed to permit only a single value for each assignment. In short, at this extreme point, the procedure becomes entirely deterministic. In our data, we typically observe very low entropy for $p(\v{\mask}|\{ \v{s}^{(k)} \})$, and so we likely operate in a regime close to the deterministic one, where the approximation is very accurate. 

\end{document}